\documentclass[manuscript,screen,review=false]{acmart}

\AtBeginDocument{%
  \providecommand\BibTeX{{%
    \normalfont B\kern-0.5em{\scshape i\kern-0.25em b}\kern-0.8em\TeX}}}

\begin{document}

\title{Named Entity Resolution in Personal Knowledge Graphs}

\author{Mayank Kejriwal}
\authornote{This preprint will appear as a book chapter by the same name in an upcoming (Oct. 2023) book `Personal Knowledge Graphs (PKGs): Methodology, tools and applications' edited by Tiwari et al. Research underlying any aspects of this work was originally conducted by the author when he was a PhD student at the University of Texas at Austin, and has been further elaborated in his dissertation \cite{kejriwal2016dis}.}
\email{kejriwal@isi.edu}
\affiliation{%
  \institution{University of Southern California}
  \streetaddress{Information Sciences Institute}
  \city{Marina del Rey}
  \state{California}
  \country{USA}
  \postcode{90292}
}








\renewcommand{\shortauthors}{Kejriwal}

\begin{abstract}
  Entity Resolution (ER) is the problem of determining when two entities refer to the same underlying entity. The problem has been studied for over 50 years, and most recently, has taken on new importance in an era of large, heterogeneous ‘knowledge graphs’ published on the Web and used widely in domains as wide ranging as social media, e-commerce and search. This chapter will discuss the specific problem of named ER in the context of personal knowledge graphs (PKGs). We begin with a formal definition of the problem, and the components necessary for doing high-quality and efficient ER. We also discuss some challenges that are expected to arise for Web-scale data. Next, we provide a brief literature review, with a special focus on how existing techniques can potentially apply to PKGs. We conclude the chapter by covering some applications, as well as promising directions for future research. 
\end{abstract}





\maketitle

\section{Introduction}

Due to the growth of large amounts of heterogeneous data (especially involving people) on the Web and in enterprise, the problem often arises as to when two pieces of information describing two entities are, in fact, describing the same \emph{underlying} entity. For instance, a company may make an acquisition and attempt to merge the acquired company's database with their own database. It is probable that there is overlap between the two, and that the acquiring company shares some customers with the acquired company. Similarly, a data aggregator may be attempting to merge together profiles from different public websites, or even social media platforms. Such platforms have heavy overlap, leading naturally to the problem of `resolving' entities. 

\begin{figure}
\centering
\includegraphics[height=5cm]{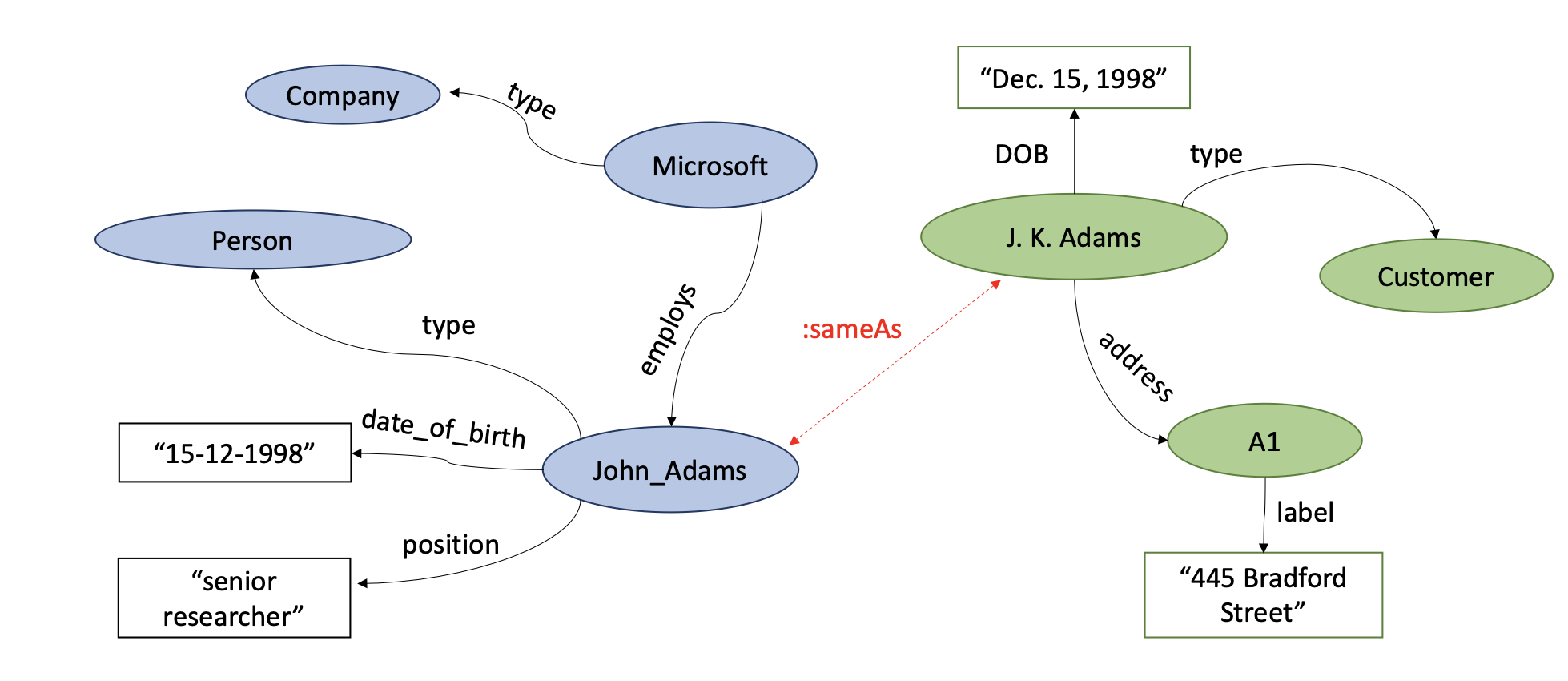}
\caption{An illustration of the named entity resolution problem between fragments of two personal knowledge graphs (PKGs).}\label{fig1}
\end{figure}


This problem, called Entity Resolution (ER), has been researched for at least half a century in text, database, and more recently, machine learning, communities, using several methodologies (e.g. rule-based vs. statistical approaches) \cite{newcombe1959a,elmagarmid2007a}. Figure \ref{fig1} illustrates a prototypical example of such `duplicate' entities that arise in two personal knowledge graphs (PKGs). An exhaustive treatment of this research is beyond the scope of this chapter, although a book level treatment has been provided by various authors, including \cite{KGbook1}, \cite{KGbook2}, \cite{KGbook3}, \cite{KGbook4}. Instead, we consider two goals in summarizing the literature on ER. First, we aim to synthesize common trends that have emerged over the last half-century. Impressively, despite much independent research across various fields and applications, there is robust consensus on several issues, including the abstract workflow of an ER implementation. Second, and in contrast with the first goal, we aim to discuss the key differences that have also emerged from this body of research. As will be subsequently discussed, many of these differences tend to be algorithmic, rather than conceptual, and are a consequence of the natural evolution of the field over time.

Prior to discussing ER itself, an important prerequisite is deciding the data model for representing the PKG. Although several options exist, we assume the primary data model to be the structured Resource Description Framework (RDF) model \cite{klyne2006a}, which is prominent in the Semantic Web. An alternative model, for datasets that are highly structured, regular and tabular in nature (which are unlikely for KG applications and domains) is the Relational Database (RDB) model. This model is important for historical reasons, given that much of the ER literature has traditionally been within the database community \cite{elmagarmid2007a}. Indeed, there are also cases in the literature where research in the RDB community has been leveraged to solve a compatible problem (e.g. query optimization) on RDF knowledge graphs \cite{angles2005a,sequeda2013a,sahoo2009a}. Hence, there is good synergy between the two models, allowing us to limit much of our treatment to the RDF model (and others similar to it) without necessarily losing generality. For the sake of completeness, we formally define an RDF graph by first defining an RDF triple:

\begin{definition}[RDF Triple]
 Given three disjoint sets of I,B and L, of Internationalized Resource Identifiers (IRIs), abstract identifiers and literals respectively, a triple in the Resource Description Framework (RDF) data model is a 3-element tuple (subject, property, object), where subject $\in$ I $\cup$ B, property $\in$ I and object $\in$ I $\cup$ B $\cup$ L. The triple is referred to as an RDF triple.
\end{definition}

Given this definition, an RDF graph can be defined as a set of triples. Visually, the literals and IRIs represent nodes in the KG (which is a directed, labeled graph by definition), while the triple itself symbolizes an edge. Note that literals cannot have outgoing edges in this model. In practice, due to Semantic Web norms, such as the four Linked Data principles \cite{bizer2009a}, IRIs in RDF KGs are just Uniform Resource Identifiers (URIs), a strict subset of IRIs. This is explicitly required by the first Linked Data principle. Furthermore, abstract identifiers (``blank nodes'') are not used in KGs that are intended to be published as Linked Data. Since PKGs, especially those acquired or constructed at large scales, and meant to be linked to other sources, are likely to obey the Linked Data principles (at least approximately), these assumptions almost always hold in practical settings.  

While RDF is the dominant data model used for representing KGs in the Semantic Web, it also has other important uses. For example, it is the basis in the full Semantic Web technology stack for representing RDF Schema (RDFS), and the Web Ontology Language (OWL) \cite{allemang2011a}. These semantic markup languages are important for publishing detailed data schemas and ontologies \cite{mcguinness2004a}, which serve as the representational metadata for the underlying KGs.

With the assumptions about the data model in place, Named Entity Resolution, henceforth called ER, can be formally defined below:

\begin{definition}[Named Entity Resolution]
 Given an RDF knowledge graph, Named Entity Resolution is defined as the algorithmic problem of determining the pairs of instances (subjects and URI objects) in the graph that refer to the same underlying entity.
\end{definition}

Entities that need to be resolved but that are not `named' tend to arise most often in the natural language setting, rather than in KG applications. In the Natural Language Processing (NLP) community, for example, anaphora and co-reference resolution are the related problems of resolving pronouns and non-named entities to their named equivalents \cite{coref1}, \cite{coref2}. Hence, such non-named mentions are not retained in the actual KG that is constructed over the raw text. Therefore, in practice, named ER is virtually identical to ER in the literature. 

A critical point to note here is the notion of an \emph{instance} in an RDF PKG. For instance, would we consider both literals and URIs to be instances? In general, the approaches that we consider for practical ER assume that an instance must have a representation using a URI, although many URIs may be disregarded from serving as valid inputs to an ER system. For example, based on the specifics of our domain-specific application, we may decide that we only want to resolve instances of `customers', rather than `contractors.' If this is the case, then URLs corresponding to contractor-entities would not be in the named entity sets input to the ER system. 

A pragmatic reason for not considering literals explicitly is that, if an entity is represented only as a literal (such as a string or a number) and has no other information or attributes associated with it, domain-specific `matching' functions would be more appropriate rather than an advanced ER solution. Several such functions are available and widely used for common attributes, such as dates, names, and addresses \cite{matchingfunc1}, \cite{matchingfunc2}, \cite{matchingfunc3}. For example, if the problem was merely restricted to matching people’s names, without any other surrounding context or attributes, a string-matching algorithm, some of which rely on phonetics \cite{pinto2012a}, could be used. We subsequently provide more details on such matching functions because, aside from being useful for matching simple literals, they are also useful for converting pairs of entity representations into numeric feature vectors. 

\section{Two-Step Framework for Efficient Named Entity Resolution}

Even in early research, the quadratic complexity of pairwise ER was becoming well recognized \cite{newcombe1959a}.  Given two RDF graphs $D_1$ and $D_2$ that we represent equivalently, with slight abuse of terminology, as sets of named entities (i.e., ignoring edges, literals, and entity-types that are not of interest), a naïve ER must evaluate all possible entity pairs, namely the cross-product $D_1 \times D_2$. Even assuming constant cost per evaluation, which may not always be the case, the run-time complexity is $O(|D_1 ||D_2 |)$. 

In the remainder of this section, for two input-sets $D_1$ and $D_2$, a named entity pair $(e_1,e_2)$ is denoted as bilateral iff $e_1  \in D_1$ and $e_2  \in D_2$. Given a collection of entities from $D_1 \cup D_2$, two entities $e_1$ and $e_2$  are bilaterally paired iff $(e_1,e_2)$  is bilateral.

\begin{figure}
\centering
\includegraphics[width=\textwidth]{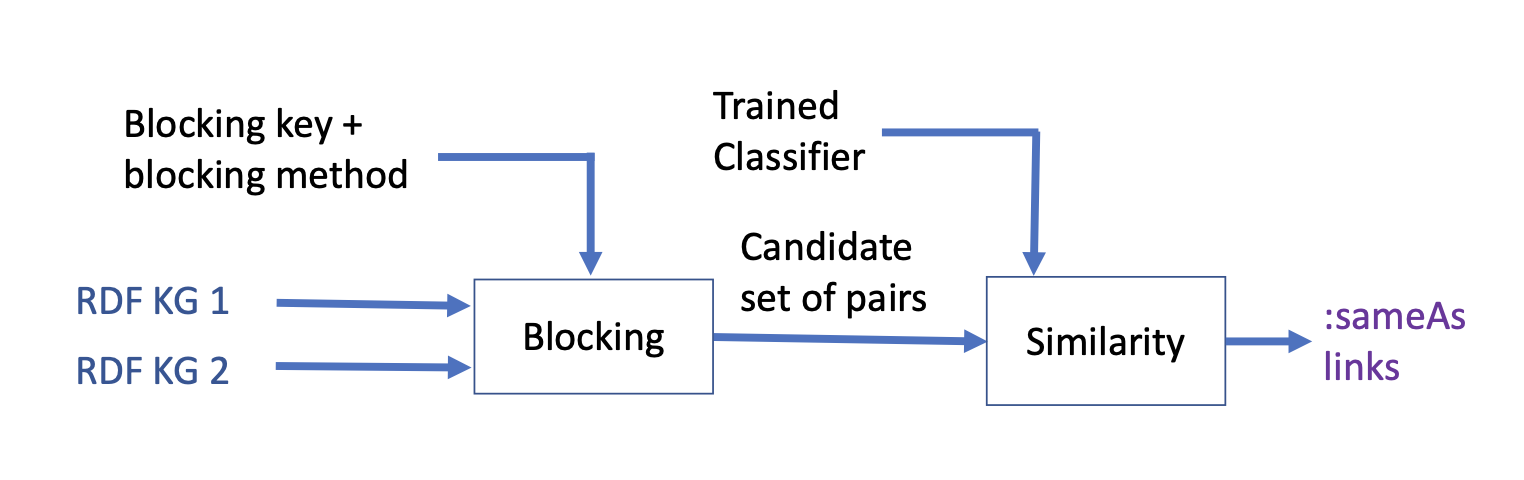}
\caption{A typical two-step ER workflow that is often implemented in practice for solving the problem efficiently and effectively. Although the figure assumes two RDF KGs, a similar workflow would also apply to non-KG datasets, as well as to the deduplication problem (where a single dataset is input).}\label{fig2}
\end{figure}

To alleviate the quadratic complexity of evaluating all possible bilateral pairs, a two-step framework is typically adopted in much of the ER literature \cite{christenbook}. This two-step framework is illustrated in Figure \ref{fig2}. The first step, blocking, uses a function called a blocking key to cluster approximately similar entities into (potentially overlapping) blocks \cite{christenblocking}. A blocking method then considers which entities sharing a block to bilaterally pair, with the result that those entity pairs become candidates for further evaluation by a matching or similarity function in the second step \cite{volz2009a}. This function, which is also called a link specification function in the literature, may be Boolean or probabilistic, and makes a prediction about whether an input entity-pair represents the same underlying entity. Prediction quality metrics, such as precision and recall, can then be used to evaluate the performance of the entire ER system by comparing the predicted matches to those present in an external ground-truth of `gold-standard' matches.

In most ER systems, $D_1$ and $D_2$ are assumed to be structurally homogeneous \cite{elmagarmid2007a, christenbook}, a term introduced in an influential survey \cite{elmagarmid2007a}. That is, $D_1$ and $D_2$   are assumed to contain entities of the same type (e.g., Person), and are described by the same property schema. The latter implies that the same sets of attribute-types are associated with entities in both data sources. An important special application of structural homogeneity is deduplication, whereby matching entities in a single data source must be found. In the rest of this chapter, we assume structural homogeneity as well, except where explicitly indicated. In practice, this assumption is not problematic because, even in the rare case of matching entities between PKGs with drastically different ontologies, an ontology matching solution can be applied as a first step to homogenize the data sources \cite{OM1, OM2, kejriwalsemantic1, tian2014a}.

\subsection{Blocking Step}

Following the intuitions described earlier, a blocking key can be defined as follows. 

\begin{definition}[Blocking Key]
Given a set $D$ of entities, a blocking key $K$ is function that takes an entity from $D$ as input and returns a non-empty set of literals, referred to as the blocking key values (BKVs) of the entity.
\end{definition}

Let $K(e)$ denote the set of BKVs assigned to the entity $e \in D$ by the blocking key $K$. Given two data sources $D_1$ and $D_2$, two blocking keys $K_1$ and $K_2$ can be defined using the definition above. Multiple definitions are typically used only when $D_1$ and $D_2$ are heterogeneous. Since we are assuming structural homogeneity in this chapter, a single key (namely, $_1=K_2=K$), applicable to both $D_1$ and $D_2$, is assumed. Without loss of generality, the literals defined above are assumed to be strings although, in principle, any data type could be used.

\begin{example}[Example of Blocking Key]
Earlier, Figure \ref{fig1} had illustrated two RDF PKG fragments describing people. Although the graphs are structurally heterogeneous, let us assume that an ontology or schema matching step had been applied such that the schema of the second graph is appropriately aligned with that of the first (e.g., DOB is mapped to date\_of\_birth, and so on). An example of a good blocking key K applicable to this schema might now be $K=Tokens(:instance) \cup Year(:date\_of\_birth)$. We assume that there is a mechanistic function that can extract the year from the date of birth literal, and that `:instance' represents the mnemonic representation (typically a label in the RDF graph) of the instance. Applied to an entity $e$ from either dataset, K would return a set of BKVs that contains the tokens in an entity’s label, as well as a single number for the year of birth, converted to a string. For example, when applied to the instance \emph{John\_Adams} from the first KG in Figure \ref{fig1}, the output (set of BKVs) returned by the blocking key would be \{``John'', ``Adams'', ``1998''\}. Similarly, when applied to the instance \emph{J. K. Adams} from the second KG, the output returned would be the set \{``J'', ``K'', ``Adams'', ``1998''\}. Since the two BKV sets have at least one common BKV, the two instances would share at least one block.
\end{example}

Given the single blocking key $K$, a candidate set $C$ of bilateral entity pairs can be generated by a blocking method using the BKVs of the entities. We briefly describe some prominent blocking methods below.

\begin{figure}
\centering
\includegraphics[width=\textwidth]{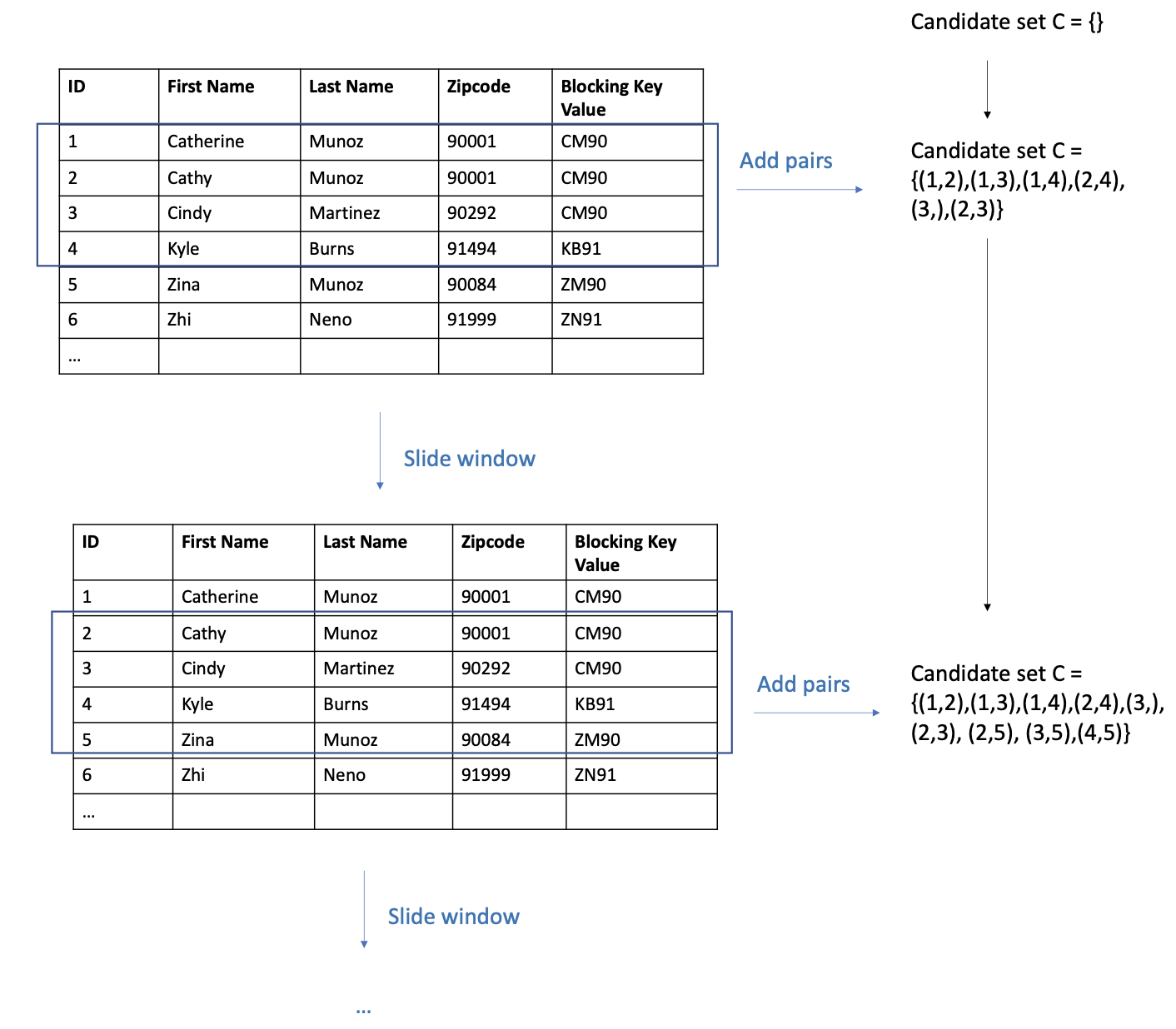}
\caption{An illustration of the Sorted Neighborhood blocking method.}\label{fig3}
\end{figure}

\begin{example}[Sorted Neighborhood]
Figure \ref{fig3} illustrates the Sorted Neighborhood blocking method on a small relational database describing people. A single BKV is first generated (in this case, we concatenate the initials of first and last-name tokens, and the first two digits of the zipcode) for each instance in the table. Next, the BKVs are used as `sorting keys'. Finally, a sliding window is slid over the table and all records within the window are paired and added to the candidate set of pairs that is input to the similarity step. For example, assuming a sliding window of 4, record pairs (1,2),(1,3), (1,4), (2,4), (3,4) and (2,3) are added to the candidate set C in the first sliding iteration, because the records with IDs 1, 2, 3 and 4 fall within the first window. The window slides forward by one record, and in the second iteration, new record pairs (2,5), (3,5) and (4,5) are added to C. The method terminates when the window cannot slide any further.
\end{example}

Besides Sorted Neighborhood and simpler blocking methods, such as simple indexing (called `traditional blocking') clustering methods, such as Canopies, have also been successfully used for blocking \cite{mccallumCC,baxter2003a}. The basic algorithm takes a distance function and two threshold parameters tight $\geq 0$ and $loose \geq tight$. It operates in the following way for deduplication (where duplicate entities must be detected in a single data source):
\begin{enumerate}
    \item A `seed' entity is chosen randomly from the set of all entities in the PKG. Let us denote this entity as $e$. 
    \item A linear-time distance-based search is conducted in the feature-representation space and all other entities in the PKG that have distance less than $loose$ to $e$ are placed into a `canopy' represented by $e$.
    \item Using the results from the above search, entities with distance less than $tight$ to $e$ are removed from further consideration. 
    \item The three steps above are repeated until each entity in the PKG has been assigned to at least one canopy.
\end{enumerate}

In the worse case, there could be entities that represent a singleton-canopy i.e., no other entities have been assigned to them. An important point to note in the above steps is that entities that have distance less than $tight$ to $e$ \emph{also} have distance less than $loose$ by definition. Thus, before being removed from further consideration, they are guaranteed to have been assigned to at least one canopy (represented by $e$). Indeed, not including the extreme (and usually, rare) case of singleton canopies, the standard behavior for most entities is that they will be assigned to at least one canopy. 

Furthermore, it is easy to extend the method to the two-PKG (or even multi-PKG) case by using the entities from the smallest PKG (or some other such well-defined selection rule) exclusively as seed entities. This extension has the added advantage of rendering the algorithm deterministic, since the randomization inherent in the original version above (which determines both the set and order of seed entities ultimately considered) is no longer present. We note that the theoretical or empirical properties of the Canopies algorithm, similar to other blocking algorithms, has not been well explored in the multi-PKG case where more than two PKGs have to be resolved. Indeed, proper algorithmic architectures for multi-PKG ER remains an under-addressed problem in the AI and database literature. 



Unlike an algorithm such as Sorted Neighborhood, the Canopies method is not actually dependent  on an `explicit' blocking key or scheme. However, that does not mean it is assumption-free. The choice of the feature space, and the distance function used, are both important decisions that serve as proxies for the blocking key required by more traditional approaches preceding Canopies. Hence, at least one paper in the literature \cite{ma2013} has referred to methods like Canopies as \emph{instance-based blocking} as opposed to \emph{feature-based blocking}, of which both Sorted Neighborhood and traditional blocking are paradigmatic examples.

However, this is not to imply that Canopies and Sorted Neighborhood do not share complementary features, both in their usage and in their developmental history. Both have been enormously popular in the ER community, and variants and versions of both, each professing to be beneficial for different use cases and datasets, have been described over the decades. That being said, the original version continues to be heavily used \cite{christenblocking}, and is a popular choice when the distance function is well defined (which is often the case for text data, since functions such as Jaccard or Cosine can be efficiently used \cite{baxter2003a}) or when it is very efficient to execute (such as certain distance functions in Euclidean space). 

One example of a variant is the use of a nearest-neighbors approach, rather than a thresholding approach, for determining which entities to assign to a canopy and to remove from further consideration. Another variant, which shares chaarcteristics of a feature-based blocking method, is to use an explicit blocking key to first generate BKVs for each entity. Canopies is then applied, not to the entities themselves, but to the BKVs of the entities \cite{christenblocking}. If more than one BKV is possible per entity, a set-based function is required.  Thus, this method is considerably more complex than the original method, but may have advantages in specific use cases.

What is the relationship between canopies and blocks? As might be evident from the terminology itself, a canopy itself can be thought of as a block. Because canopy `blocks' can be overlapping, the blocks are different from those found in traditional blocking where it is common to assume that blocks are not overlapping (i.e., that nodes can be assigned to at most one block, or that BKV sets always have at most size one). 


Some techniques have been proposed in both the ER literature (but also beyond) that could potentially be considered as `alternatives' to blocking. A method that is especially noteworthy in this regard is \emph{Locality Sensitive Hashing (LSH)}. In recent times, LSH has become popular in the Big Data community for presenting an efficient (albeit, approximate) solution to the important problem of nearest-neighbors search in spaces that have high dimensionality \cite{datarLSH}. Specifically, given a  an LSH family of hash functions may be defined using five parameters: (i) a distance measure $d$, (ii) two `radii' denoted as $r$ and $s$, with $r<s$, (iii) two probabilities ($p_r$, $p_s$ with $p_r> p_s$) associated with $r$ and $s$ as suggested by the terminology. 

An LSH family is considered to be $(r,s,p_r,p_s)$-sensitive if it is the case that any point $u$ falling inside a sphere with radius $r$ centered at point $v$ (note that the computation of this obviously depends on the feature space and $d$) necessarily has the same hash as $v$ with a minimum probability of $p_r$, dependent on a probability distribution defined on the hash family. Furthermore, it is also required that if $u$ does not fall within a similarly defined sphere, but with radius $s$, the probability that $u$ and $v$ have the same hash is at most $p_s$ \cite{datarLSH}. Such $(r,s,p_r,p_s)$-sensitive families are widely used in practice, and a number of them have been defined in recent papers, because highly efficient algorithms can be designed to hash big datasets and determine (with some probability) when two points are very close to (or far from) each other. 

One can see the application of this algorithm to blocking under the right conditions, as the goal of blocking is very similar to that of LSH. Indeed, an LSH-like algorithm is directly applicable to instance-based blocking methods like Canopies that rely explicitly on distance functions rather than blocking keys. However, it must be borne in mind that the requirements of LSH are stricter, whereas Canopies can be used with any well-defined distance measure. A good example of a distance function which is amenable to LSH is the Jaccard function, which can be approximated through the MinHash function  (which has the requisite properties and sensitivity defined earlier). Certainly, despite its rigid assumptions, LSH should be considered as a viable baseline for more complex blocking methods. Beyond ER, it has also been applied to problems such as ontology alignment \cite{duan2012instance}, and on occasion, it has also been applied to the similarity step of ER. 


\subsection{Similarity / Matching Step}

Although the expectation is that the candidate set generated by blocking contains most (and in the ideal situation, all) of the true positives (duplicate entity pairs) present in the PKGs being resolved, its approximate and efficient nature also leads to many true negatives being present\footnote{Interestingly, the issue of false positives does not arise, because the similarity step has to make the final decision on what constitutes a positive. Hence, it does not make sense to determine which pairs are `false' positives since one similarity function may falsely declare a pair to be a duplicate (hence, false positive) while another may not. True positives are a subset of the pairs in the ground-truth set of duplicates and must be included in blocking for \emph{any} similarity function to make that determination.}. Additionally, a few false negatives may get `excluded' from the candidate set. In a subsequent section, we discuss the evaluation of blocking to measure the extent to which this is taking place. Note that, once excluded, there is no hope of recovering `false negatives' (except to re-execute a different blocking algorithm or two-step workflow on the entire dataset) in the similarity step. However, a `good' similarity function can distinguish between true positives and true negatives effectively, leading to better results in the overall two-step workflow.  Indeed, it is easy to see that the similarity function must be finer-grained, compared to blocking, as it needs to make that distinction, which blocking proved incapable of doing due to its focus on efficiency rather than effectiveness \cite{elmagarmid2007a}. 

While the similarity function is referred to differently in different communities, within the Semantic Web, it is often referred to as a \emph{link specification} function \cite{volz2009a}. We adopt similar terminology here due to the chapter's focus on PKGs rather than relational databases or other similar data models. 

\begin{definition}[Link Specification Function]
Given two data sources $D_1$  and $D_2$, a link specification function is a Boolean function that takes as input a bilateral pair of entities, and returns True iff the input entity pair refers to the same underlying entity (i.e., is a duplicate pair) and returns False otherwise.
\end{definition}

Because the underlying real-world link specification function, if it even exists analytically, is not known or discoverable given small sets of duplicate examples, it has to be inferred or learned. In some cases, domain knowledge is used to devise sets of rules as a viable link specification function. In all but the most trivial cases, the function is approximate and unlikely to be perfect. In the treatment below, we assume the `link specification function'  to mean the approximated version of the function rather than the unknown underlying function. Note that the function can be real-valued i.e., given a pair of entities, it may return a value in $[0,1]$, which can be properly interpreted as a score or the model's belief that the pair represents a duplicate. It is also not uncommon to have `hardened' link specification functions that would just return 0 or 1.

With the preliminaries above, the similarity step in the two-step workflow can be thought of as the application of $L$ to each pair in the candidate set \cite{elmagarmid2007a,christenbook}. If $L$ is real-valued, further assumptions are required to make the distinction between duplicates and non-duplicates. A simple approach is to use a threshold (determined heuristically, or through a set of `development set' duplicates and non-duplicates set aside for exactly this purpose): pairs in the candidate set for which $L$ returns a score greater than the threshold are classified as duplicates and similarly for non-duplicates. In theory, one can even have two thresholds (denoted as $lower$ and $upper$), not dissimilar to blocking methods like Canopies. Pairs with scores that fall between the thresholds are usually considered to be indeterminate and flagged for manual review, or (in more complex systems) undergo processing by a more expensive, and separately trained, link specification function. 

Many link specification functions have been explored by researchers through the decades, including in the expert systems community, machine learning, natural language processing, Semantic Web, and databases. For a good survey of such techniques in databases (primarily), we refer the reader to \cite{elmagarmid2007a}, as well as a book on data matching \cite{christenbook}. Below, we provide more details on a machine learning-centric approach to the problem. 


\begin{figure}
\centering
\includegraphics[width=\textwidth]{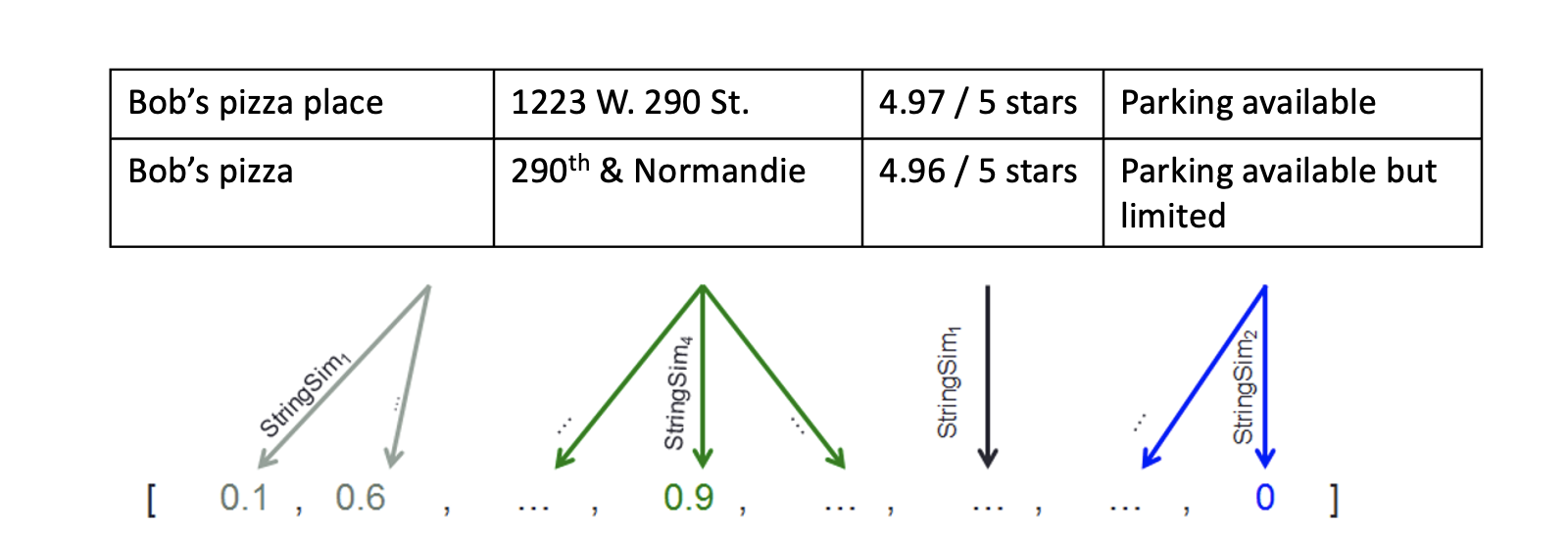}
\caption{Conversion of an entity pair to a feature vector that could then be used in ordinary classification systems (e.g., deep neural networks).}\label{fig4}
\end{figure}

If using machine learning to derive the function $L$, the most important step is to determine how to extract a feature vector (whether real-valued or discrete) given a \emph{pair} of entities. The procedure is illustrated schematically in Figure \ref{fig4}, for the case where both entities (represented as records) share the same schema or ontology. Recall that we referred to this common case as one exhibiting \emph{structural homogeneity}. As the figure illustrates, a library of feature functions is assumed. The feature function is like a `primitive': it takes a pair of primitive data types (such as strings or dates) as input and returns a numeric feature. Without loss of generality, let us assume that the output is always real-valued. Given such a library of feature functions, a pair in the candidate set can be converted to a well-defined and fixed-length feature functions. Indeed, if we assume $m$ feature functions, and $n$ such `fields' representing the schema of the entity, the feature vector would have $mn$ dimensions\footnote{Note that this could also be the case when an entity is missing a value for a given field, or if the feature function fails for some reason. Such cases could be encoded by using a special value (such as -1) in its position in the feature vector. Such preprocessing is important, and largely a function of robust engineering, and ensure that the (ordered) feature vector always has $mn$ well-defined values. Variable length feature vectors are typically very difficult for the majority of machine learning algorithms to handle.}. 

Fortunately, many feature functions are available to practitioners. Some have been known for almost a century, while others are more recent. Primitive data types processed by these functions include strings, tokens, and even numbers. In some cases, the semantics of the field can play a role. For instance, \emph{phonetic} feature functions are an excellent fit for strings that represent names. Good descriptions may be found in a variety of sources, including \cite{elmagarmid2007a} and \cite{christenbook}. Domain-specific feature functions have also been proposed in the literature, but may be more difficult to locate or implement. Interestingly, neural networks have been used frequently in recent years to bypass the application of hand-picked feature functions. For example, the work in \cite{kejriwalGeonames} considers the use of such representation learning for extracting features from geographically situated entities represented using latitudes and longitudes, rather than strings or other descriptive information. 

An alternative way of extracting features is by generating hashes using several well-known LSH families \cite{kejriwal2015c}. According to this model of feature generation, the underlying link specification function can be modeled through a functional combination of various distance measures for which LSH-sensitive families exist. A validation of this model would be consequential as it significantly eases the burden of scalability, both in the blocking and similarity steps. In the most general case, the hashes could be used as features, and an appropriate learner would be used for discovering an explicit functional combination (or rules) for class separation. The process and evaluation are described in more detail in Chapter 7 of Kejriwal (2016).  

A machine learning classifier is trained on positively and negatively labeled training samples, and is used to classify vectors in the candidate set. Several classifiers have been explored in the literature, with random forest, multilayer perceptron and Support Vector Machine classifiers all found to perform reasonably well \cite{rong2012, soru2014, kejriwal2015c}. In more recent years, neural networks have also been applied \cite{ERnn}.

\subsection{Independence of the Blocking and Similarity Steps}

The two-step workflow, and our description of blocking and similarity, seems to suggest that the two steps are largely independent. Historically, and even in actual practice, this has been the case, and it is even possible to `swap' out different blocking algorithms while keeping the similarity step constant (and similarly for swapping the similarity modules while keeping blocking constant). However, it behooves us to mention that nothing \emph{prevents} blocking and similarity from being interlocked. Namely, the two steps can be set up such that they `interact' in a real system i.e., we do not need the candidate set to be completely generated prior to executing the similarity step. A simple example of this is a system that tries to conserve space by not storing the candidate set explicitly, but piping pairs to the similarity step \emph{as} they are generated by blocking. The system still maintains the assumption, however, that decisions made in the similarity step have no impact on blocking (i.e., there is no backward feedback). Furthermore, depending on the blocking method used, there may be a loss in overall efficiency, since the same pair may be classified several times by the similarity step (as we are not maintaining a set-based data structure). Even in this simple example, the tradeoffs between time and space efficiency must be carefully negotiated. 

A small, but growing, number of applications in recent years have also been challenging the backward-feedback assumption we mentioned above, example references being \cite{whang2009a, papadakis2013a}. An example of a blocking method that takes backward feedback into account is  \emph{comparisons propagation}, which tries to use decisions made in the similarity step to \emph{estimate}, in real time, the \emph{utility} of a block \cite{papadakis2012blocking}. Intuitively, if a block is yielding too many non-duplicates (according to the similarity function), then it may be best to cut losses and stop processing the block rather than continue processing it and generating more `useless' pairs for the similarity step to classify as non-duplicates. In a rational decision-making setting, the expected gain from processing the block any further is outweighted by the loss in efficiency, making it appropriate to discard it. 

While such techniques are promising, their implementations have mostly been limited to serial architectures, owing to the need for continuous data-sharing between the similarity and block generating components \cite{whang2009a, papadakis2012blocking}. Experimentally, the benefits of such techniques over independent techniques like Sorted Neighborhood or traditional blocking (with skew-eliminating measures such as block purging) have not been established extensively enough to warrant widespread adoption. Therefore, the two-step workflow, with both steps relatively independent, continues to be predominant in much of the ER research \cite{kopcke2010a}.

\section{Evaluating Named Entity Resolution}

The independence of blocking and similarity suggests that the performance of each can be controlled for the other in experiments \cite{elmagarmid2007a}. In the last decade both blocking and similarity have become increasingly sophisticated. It is now the rule, rather than the exception, to publish either on blocking or on similarity within an individual publication \cite{christenblocking}. Despite some potential disadvantages, this methodology has yielded the adoption of well-defined evaluation metrics and standards for both steps.  

\subsection{Evaluating Blocking}

The primary aim of blocking is to scale the naïve similarity-only ER system that bilaterally pairs all entities with one another. Blocking serves this goal by generating a smaller candidate set than this `exhaustive set'. However, if time-complexity reduction were the only goal, optimal blocking would simply yield the empty set as the candidate set. Such a system would ultimately be without utility because it would generate a candidate set with zero duplicates coverage.

In other words, duplicates coverage and candidate set reduction are two competing goals that every blocking technique (blocking key and blocking method) aims to trade off between \cite{hernandez1998}. This tradeoff can also be formalized. As a first step, let us denote $\Omega$ as the set $D_1 \times D_2$; in other words, the exhaustive set of all bilateral pairs. Let $\Omega_M$ denote the subset of $\Omega$ that contains all (and only) matching entity pairs. $\Omega_M$ is designated as the ground-truth (equivalently, gold standard). Finally, we use C to denote the candidate set generated by applying the blocking method. Using this notation, Reduction Ratio (RR) can be defined using the formula below:

\begin{equation}
    RR = 1-\frac{|C|}{|\Omega|}
\end{equation}

A higher RR implies greater time-complexity reduction (and depending on the infrastructure, space-complexity reduction) achieved by the blocking method, compared to generating the exhaustive set \cite{christenblocking}. Although less commonly used for this purpose, RR can also be evaluated relative to the candidate set of a baseline blocking method \cite{papadakis2013a}. The only change required in the formula above is to replace $\Omega$ with the candidate set of the baseline method. In its relative usage, such an RR would have positive value if the blocking method resulted in greater savings compared to the baseline method; otherwise, the relative RR would be negative. 

Importantly, even minor differences in RR can have an enormous impact in terms of run-time because of its quadratic nature. For instance, consider the case where $\Omega$ contains 10 million pairs (a not unreasonable number that could be achieved if the two data sources only had around 1,000-2,000 instances each). An improvement of even 0.1\% on the RR metric would imply savings of thousands of pairs. The same percentage could represent millions of pairs on even larger datasets.

While RR is a good way of measuring the complexity reduction goal of blocking, the Pairs Completeness (PC) metric,  defined below, quantifies the method’s duplicates coverage:

\begin{equation}
    PC = \frac{|C \cap \Omega_M|}{|\Omega_M|}
\end{equation}
	
Interestingly, the PC serves as an upper bound on the recall metric that is used for evaluating overall duplicates coverage in the ER system (i.e., after the similarity step has been applied on the candidate set). For example, if PC is only 70\%, meaning that 30\% of the duplicate pairs did not get included in the candidate set, then the ER system’s overall recall can never exceed 70\%. 

While not theoretically necessary, in practice, there is a tradeoff between achieving both high PC and RR. In most blocking architectures, the tradeoff is negotiated by tuning one or more relevant parameters. For example, the sliding window parameter w in Sorted Neighborhood (illustrated earlier through an example) can be increased to achieve higher PC, at the cost of lower RR \cite{hernandez1998}. 

Although we can always plot PC-RR curves to visually demonstrate the tradeoff in a blocking system, a single number is desired in practical settings. In the literature, this number is usually just the F-Measure, or harmonic mean, between the PC and RR:

\begin{equation}
    F-Measure = \frac{2.PC.RR}{PC+RR}
\end{equation}

A second tradeoff metric, Pairs Quality (PQ), is less commonly known than the F-Measure of PC and RR in the wider community, but can be illuminating for comparing different blocking architectures:

\begin{equation}
    PQ = \frac{|C \cap \Omega_M|}{|C|}
\end{equation}

Theoretically, PQ may be a better measure of the tradeoff between PC and RR than the F-Measure estimate, which weighs RR and PC equally, despite the quadratic dependence of the former. For this reason, PQ has sometimes been described as a precision metric for blocking \cite{christenblocking}, although the terminology is superficial at best. The intuitive interpretation of a high PQ is that the generated blocks (and by virtue, the candidate set C) are dense in duplicate pairs. A blocking method with high PQ is therefore expected to have greater utility, although it may not necessarily lead to high precision or recall from the overall ER architecture. 

PQ can sometimes give estimates that are difficult to interpret. For example, suppose there were 1,000 duplicates in the ground-truth, and the candidate set C only contained ten pairs, of which eight represent duplicates. In this example, PQ would be 80\%. Assuming that the exhaustive set is large enough, such that RR is near-perfect, the F-Measure would still be less than 2\% (since PC is less than 1\%). The F-Measure result would be correctly interpreted as an indication that, for practical purposes, the blocking process has failed. The result indicated by PQ alone is clearly misleading, suggesting that, as a tradeoff measure, PQ should not be substituted for the F-Measure of PC and RR. An alternative, proposed by at least one author but not used widely, is to compute and report the F-Measure of PQ and PC, instead of PC and RR \cite{christenblocking}.  

\subsection{Evaluating Similarity}

As similarity is most similar to a machine learning classification problem (such as sentiment analysis), the best way to evaluate it given the real-world requirements of ER is to compute precision and recall. Specifically, once a candidate set $C$ is output by blocking and processed by similarity, the final expected output is a \emph{partition} of $C$ into sets $C_D$ and $C_{ND}$ of pairs of resolved (`duplicated') and non-resolved (`non-duplicates') entities.

Given a ground-truth set $\Omega_M$ (using the terminology of the previous section) of actual duplicate entity-pairs, the true positives (TPs), false negatives (FNs), and false positives (FPs) can be computed in the usual way. Precision is just the ratio of TPs to the sum of TPs and FPs, while recall is the ratio of TPs to the ratio of TPs and FNs. Like most machine learning problems, optimizing precision in an ER application can come at the cost of optimizing recall. One way to measure the tradeoff is by plotting a Receiver Operating Characteristic (ROC), which plots true positives against false positives \cite{hanley1982}. However, it is simpler to obtain a single-point estimate of this tradeoff by computing the harmonic mean (F-Measure) of precision and recall. Note that if F-Measure is used to evaluate similarity, as well as blocking (along the lines discussed in the previous section), a clear distinction must be noted between the two when reporting the results, as they are measuring tradeoffs between different quantities. An alternative to a single-point estimate is a precision-recall curve that, although related in a complex way to the ROC curve, expresses the tradeoff much more directly at different (precision, recall) points. Historically, and currently, precision-recall curves dominate ROC curves in the ER community \cite{kopcke2010a, kopcke2010b, menestrina2010a}. 



We emphasize that computing a measure such as accuracy is usually not a good idea. The reason is that the vast majority of pairs are expected to be non-duplicates, meaning that if a system were to predict every pair in the candidate set to be a non-duplicate pair, the resulting accuracy would be very high! Such a measure is clearly without utility. Precision and recall preempt this problem because neither takes true negatives into account in the computation.

\section{Evolution of Research in Named Entity Resolution}

\begin{figure}
\centering
\includegraphics[width=\textwidth]{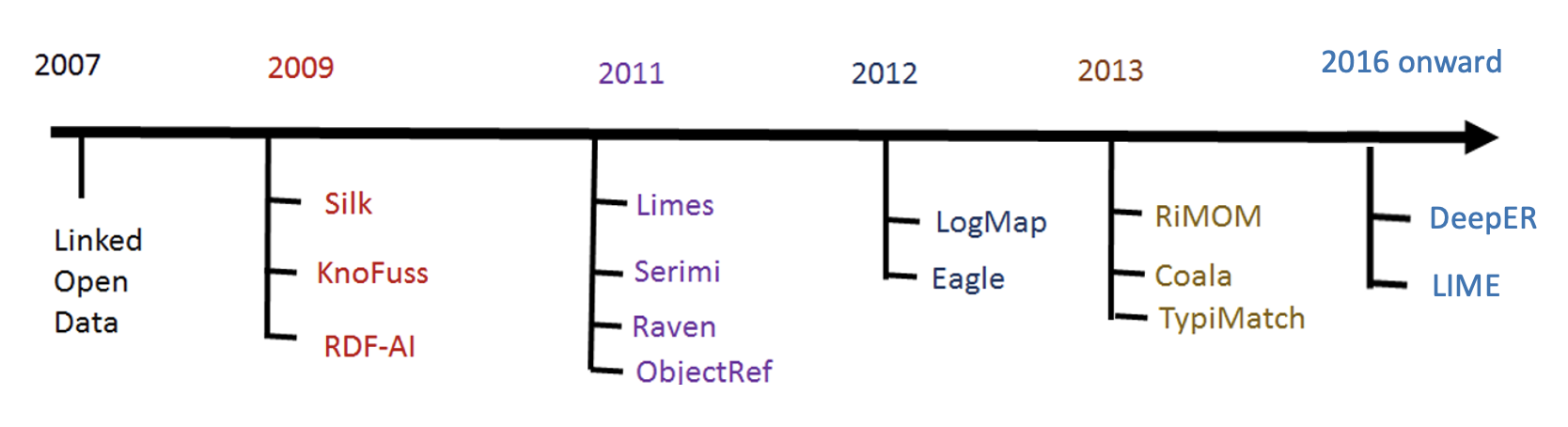}
\caption{A timeline of approaches (primarily for the similarity step) that have evolved to address ER.}\label{fig5}
\end{figure}

Owing to its 50-year history, many systems and research methods have been proposed for ER. Figure \ref{fig5} shows how different streams of thought have emerged over the decades. While necessarily simplified (e.g., many of these classes of approaches have overlapped, and some are still widely used), the figure largely tracks what has happened in the broader AI community and its many applications. For the more technical reader, we recommend books on both ER and knowledge graphs (which tend to contain chapters on ER) \cite{KGbook1, KGbook2, KGbook3, ERbook}.

In the Semantic Web community, rule-based systems were considered state-of-the-art \cite{volz2009a}, and are still widely used, in the immediate aftermath of Linked Open Data (which opened up ER as an important problem in that community). However, due to their many advantages, machine learning approaches have gained in popularity over the last decade. Such approaches, primarily supervised in nature, take training pairs of duplicates and non-duplicates and learn optimize parameters in a given hypothesis space to achieve high performance on an unseen test set. However, we note that rule-based and machine learning approaches are not necessarily exclusive or independent of one another. Indeed, a number of interesting hybrid approaches have been proposed, some of which are named in Figure \ref{fig5}. Also of interest are `low supervision' approaches that rely on techniques such as active learning \cite{ngomo2013, DBLP:conf/esws/KejriwalM15a, DBLP:conf/esws/KejriwalM15b}. Deep learning has also been applied with some success to ER \cite{ERnn}. 

Beyond the similarity step, where machine learning has been applied and studied much more extensively, methods have been proposed in the last two decades to learn blocking keys given training data of duplicates and non-duplicates \cite{BKL1, BKL2, BKL3, DBLP:journals/corr/Kejriwal16}. More recently, such approaches have also been tried for heterogeneous data collections \cite{DBLP:journals/corr/KejriwalM15}. This was quite a departure from the traditional approach, which was to manually specify blocking keys that seemed `intuitive'. As multiple approaches have now showed, a systematic approach to learning blocking keys can lead to non-trivial savings in complexity reduction, or in improved PC and PQ. The technique for learning blocking keys is less straightforward than training classifiers for similarity, but theoretically similar to problems such as set covering \cite{setcovering}. Along with blocking scheme learning, research also continues on developing new blocking methods (or variants of existing methods, like Sorted Neighborhood) for novel ecosystems such as Linked Open Data \cite{DBLP:conf/esws/KejriwalM15c}.



\section{Challenges and Opportunities for Named Entity Resolution}

Despite being a 50-year old problem, and improvements in Artificial Intelligence and deep learning technology, ER remains a challenging problem due to the rapid growth of large and heterogeneous datasets published on the Web. In the previous section, a critical challenge that was described is the non-obvious ability of existing systems to simultaneously address challenges such as domain-independence, scalability and heterogeneity. At the same time, in the existing AI literature, there are theoretical and applied mechanisms to address some of these challenges. A machine learning paradigm such as transfer learning, for example, could potentially be used to handle the domain-independence requirement by first bootstrapping an ER system in a few `anchor' domains, and then using transfer learning to adapt the system to other domains. In theory, such an approach seems feasible, but in practice, it is very difficult to execute. Some other challenges, which were also expounded upon in \cite{kejriwal2016dis} are:

\begin{enumerate}
    \item {\bf Schema-free approaches to ER:} The increased diversity and heterogeneity of PKGs published on the Web indicates that the present time is a good one for further investigating so-called \emph{schema-free approaches} to ER. We note that the traditional approaches have been primarily inspired by the Relational Database community, which tends to assume some form of schema or ontology matching in order to `align' the types of instances before processing them further (e.g., through the two-step workflow). Conventionally, many algorithms considered such a homogeneous type-structure as a given, for both qualitative and computation reasons. We already saw in the previous sections that many algorithms only seem to apply if the structural homogeneity condition is met. However, more recently, there has been much literature calling this assumption into question and seeking to extend methods to work without it. In some of our own papers, we suggested a schema-free implementation of the classic Sorted Neighborhood algorithm that is specifically designed for RDF KGs, and empirical results showed that the method compares favorably to the more established baseline \cite{kejriwalSN1, kejriwalSN2}. However, schema-free approaches to ER remain relatively novel in the community, and many conceptual and methodological questions remain. For example, how can we tell which schema-free features are of `good' quality, and construct such feature functions effectively? Can deep learning play an important role to automate such construction? Finally, could such approaches allow us to bypass ontology matching altogether prior to ER? And how do we apply such approaches to domains that are not completely structured, but contain a mix of structured data and free text (of which social media applications, augmented with user meta-data, are good examples \cite{DBLP:journals/fi/HuRKL21,DBLP:conf/ssci/KejriwalFZ21,DBLP:journals/osnm/KejriwalWLW21,DBLP:journals/snam/KejriwalZ20})?
    
    \item {\bf Improving Linked Data Quality:} We discussed earlier that an important prior step, often treated as an assumption in core ER research, to conducting Named Entity Resolution is to first determine which types of nodes between two KGs should be aligned. In the deduplication scenario (as opposed to resolving entities in two or more PKGs), this problem can also arise. For example, if an individual is both an author and professor, and these are two types in the KG, then it is plausible that two instances of the individual are present in the PKG. Only executing an ER algorithm on the set of \emph{professor} instances, or on the set of \emph{author} instances (independently) is unlikely to yield high-quality results. Hence, proper type alignment is generally necessary to find the balance between running the two-step ER workflow on the full set(s) of nodes \cite{DBLP:journals/corr/KejriwalS17a, kejriwalTA}, even with blocking, and being too restrictive in which pairs of types should be considered as `aligned' for the purposes of being eligible for such processing. 
    
    On the Web, the problem of type alignment is exacerbated because of the quality of Linked Data. Many different types of entities are present, and it is not always evident whether one type should be aligned with another. A related problem is property alignment. Because this problem particularly arises with `cross-domain' or encyclopedic KGs, such as DBpedia (which contains over 400 types), improving the quality of Linked Data published online, as well as re-using existing ontological types, is an important future direction. One reason why this continues to be challenging is that there is also ambiguity surrounding the construction of good ground truths for highly complex types. In a preliminary experimental work that we published \cite{kejriwalTA}, we found that that there may be at least three different ways (both inductive and deductive) of constructing reasonable ground truths,  which are not always consistent. This can cause an extra layer of noise when evaluating type alignment (and following that, ER) systems. There is reason to believe that such noise is not merely hypothetical but actually exists in real Linked Open Datasets currently on the Web \cite{kejriwal2016dis}. On the positive front, because there are far fewer types than entities and properties, improving their quality and enforcing better publishing standards can have lead to outsize progress in addressing these issues. In turn, adoption of Linked Data improves and more ambitious data integration applications at Web scale become feasible.

\item {\bf Transfer Learning:} Transfer learning, surveyed in \cite{pan2010} (among many other papers), is a valuable avenue to pursue for resolving entities in large Linked Open datasets without requiring enormous amounts of training data. Although transfer learning (even for ER) is not a completely novel line of research \cite{baxter1998}, it has not found widespread utility yet in ER due to several technical challenges, and relatively low quality compared to fully supervised approaches. Even in other machine learning research, its progress is not completely evident, although it bears saying that it has continued to be researched actively in several mainstream applications \cite{mesnil2012}.  However, recent advent of large language models like BERT and ChatGPT suggest that feasible transfer learning solutions may be around the corner. `Pre-trained' versions of such models, which can be downloaded off-the-shelf and fine-tuned on specific datasets at relatively low cost \cite{bert}, contain much background knowledge that could boost performance on ER datasets. At the same time, these models also have some problems (such as lack of explainability but also others \cite{DBLP:journals/corr/abs-2210-01258, DBLP:journals/corr/abs-2210-07519,DBLP:conf/sgai/0003K21,DBLP:journals/corr/abs-2011-09159}) that may pose problems in domains where a high degree of confidence is needed.
Beyond classic transfer learning, other novel learning approaches, such as zero-shot and few-shot learning, also continue to be investigated for difficult machine learning problems \cite{zsl, fsl}. Such techniques could prove to be essential for ER, especially with the advent of neural language models \cite{bert}.
    
    \item {\bf ER in PKGs versus Personal Knowledge Networks (PKNs):} Although KGs have witnessed increased adoption in multiple domains over the last ten years, there has been a similar (and largely independent) rise in the field of \emph{network science}. Network science has proven to be particularly powerful in understanding complex systems \cite{DBLP:journals/corr/abs-2203-06491}, especially those where relational structure plays an important role. A classic example is social networks, which bears close resemblance to PKGs. In recent years, it has also been applied to understanding other interesting and high-impact social domains, such as illicit finance \cite{DBLP:journals/ans/KejriwalD20}, economics \cite{DBLP:journals/corr/abs-2211-13117}, international geopolitics and humanitarian applications \cite{DBLP:journals/fdata/Kejriwal21, DBLP:conf/semweb/KejriwalPZS18}, e-commerce \cite{DBLP:journals/eaai/KejriwalSNT22, DBLP:journals/cii/KejriwalSNT21,DBLP:conf/asunam/KejriwalSNT20}, social media analytics \cite{DBLP:conf/sai/LuoK22, DBLP:journals/data/MelotteK21,DBLP:journals/corr/abs-2108-01699,DBLP:journals/corr/abs-2111-05823}, crisis informatics \cite{DBLP:conf/asunam/KejriwalZ19}, misinformation and sensationalism detection \cite{DBLP:conf/asunam/ZhangK19}, and human trafficking \cite{DBLP:journals/ans/KejriwalG20,DBLP:journals/ans/KejriwalK19, DBLP:conf/aies/HundmanGKB18}.  ER in the network domain is far less studied, as has its connections to ER in KG-centric communities like Semantic Web. Visualization of outputs, designing of better interfaces, and efficient human-in-the-loop tooling, both in network science and Semantic Web, remains under-studied as well \cite{DBLP:conf/asunam/KejriwalZ19a, DBLP:conf/sigir/KejriwalSS19, DBLP:conf/aaai/KejriwalS18, DBLP:conf/chi/KejriwalS18, DBLP:conf/www/KejriwalGSC18}. At the same time, there is clearly a vital connection, and not just because PKGs and networks can both be represented as graphs. To take the human trafficking domain as an example, KGs have also been applied to the problem with significant success \cite{DBLP:journals/tbd/KejriwalS22, DBLP:journals/simpa/Kejriwal21,DBLP:journals/fi/KejriwalS19, DBLP:journals/expert/KejriwalSK18}. Another example is e-commerce \cite{DBLP:conf/sigir/GheiniK19, DBLP:journals/corr/BalajiJKMSO16}.  
    
    \item {\bf Theoretical Progress in ER as an `AI-Complete' Problem:} Named Entity Resolution is an inherently practical problem, but a theoretical understanding of the problem could open up new algorithmic directions. An example of such a treatment is the Swoosh family of algorithms \cite{swoosh}. More recently, with an increased focus on Artificial General Intelligence (AGI), there is an open question as to whether ER can be viewed as an `AI-complete' problem i.e., solving ER with sufficient accuracy and robustness may provide concrete evidence that we have made definitive progress on AGI. Examples of AI-complete problems include commonsense reasoning \cite{DBLP:journals/natmi/KejriwalSMM22} and knowledge representation \cite{DBLP:journals/corr/abs-2210-01263}, open-world learning (including reinforcement learning techniques for open-world learning \cite{DBLP:journals/simpra/KejriwalT21, DBLP:journals/corr/abs-2103-00683}). These problems, if solved, could revolutionize the applications of AI in a variety of domain, and become the underlying basis for industries of the future  \cite{DBLP:books/sp/Kejriwal23}. However, much theoretical work needs to be done before any of these claims can be validated with certainty.
    
    
    Beyond developing better theoretical foundations, one must also bear in mind that ER does not exist in isolation, but that noise in real-world ER systems may have close connections to noise in other steps of the KG construction and refinement pipeline, such as information extraction, data acquisition \cite{DBLP:journals/semweb/KejriwalSL19,DBLP:journals/information/Kejriwal22, DBLP:conf/esws/2022text2kg} and general knowledge capture \cite{DBLP:conf/kcap/2019}. Scalable ER, especially in massive distributed ecosystems like schema.org \cite{DBLP:conf/aaai/KejriwalSNT21,DBLP:journals/corr/abs-2007-13829, DBLP:journals/computer/NamK18} or other Web-scale and `Big Data' applications \cite{DBLP:journals/aimatters/Kejriwal17,DBLP:conf/aaai/Kejriwal15a,DBLP:conf/semweb/Kejriwal14}, also remains an important under-addressed problem in the research community. This problem is only likely to get worse as advanced extensions to Linked Open Data \cite{DBLP:conf/semweb/KejriwalS19} and other such ecosystems are proposed in the years to come (in part, due to the advent of large language models). In the network science community, there has also been increased focus on algorithmic scalability \cite{DBLP:conf/www/0001VDMCKR18}.
    
    \item {\bf Other Applied Directions:} As PKGs are used in different ways in different domains and use-cases, the issue of properly building domain-specific PKGs (of which ER is an important component) remains critical and is an important application of the general research direction of domain-specific KG construction \cite{DBLP:conf/www/SzekelyK18}. Semantic search, especially in domain-specific applications \cite{DBLP:conf/bigdataconf/KejriwalLJJ15}, is another important direction as it will likely require advances in domain-specific search \cite{DBLP:conf/semweb/KejriwalSS17,DBLP:conf/semweb/KejriwalS17a}. Because of the rise of large language models, and deep learning more generally, we believe that an ER approach that melds the best of traditional approaches to the problem with deep learning could lead to significant advances in upcoming years \cite{DBLP:journals/ojsw/KejriwalS17}. This is true for KG research, more generally \cite{DBLP:conf/esws/2018dl4kgs}. An example of such an application is hypothesis generation and geopolitical forecasting using (for example) multi-modal KGs rather than directly using raw text or video sources \cite{DBLP:conf/tac/ZhangSSHLPLZWWJ18}. Finally, many applications require processing and conducting ER on datasets that are structurally heterogeneous. Even today, structural homogeneity remains a strong assumption in the ER community. Removing this assumption from future approaches is a promising direction \cite{DBLP:conf/aaai/Kejriwal15}, but is not without challenges \cite{DBLP:journals/corr/Kejriwal16,DBLP:conf/semweb/KejriwalM14}.

\end{enumerate}

\section{Conclusion}

Named Entity Resolution is an important application that has been researcher for almost half a century. While the early applications of ER were limited to patient records and census data, they have proliferated in an era of Big Data and open knowledge. It is likely to continue playing an important role in Artificial Intelligence efforts in both government and industry in the near- and long-term future, especially in data-intensive text applications. As a recent book discussed, industry is now starting to invest intensively in AI systems to grow or maintain a competitive advantage, and open assets such as Linked Data and off-the-shelf large language models are important drivers in such implementations. Because of the open nature of these assets, there is a more level playing field than was historically the case. Properly implementing, scaling, and evaluating advanced algorithms, and ensuring that the data is `clean', may well provide the edge in many cases. For most datasets, achieving such quality requires concerted effort in ER. `Properly' in this context also implies gaining a better understanding of bias in training or evaluating an ER model \cite{DBLP:conf/semweb/KejriwalM15}, and other weak spots of the model. 

This chapter provided a background on ER essentials. Particularly important is the two-step workflow, comprising blocking and similarity, and their respective evaluation methodologies and metrics, which has emerged as a standard for ER. However, many important research questions and opportunities still remain, several of which we covered in the previous section. Another important research area that is particularly important to ER but that we did not mention in the previous section is the efficient generation of unbiased training sets. While this may not prove to be a problem if zero-shot learning and few-shot learning techniques are adapted to address ER, at present, only supervised learning techniques currently have the quality necessary for high-stakes applications. Hence, generation of high-quality training sets remains an important problem. Unlike regular machine learning applications, random sampling and labeling does not work very well because most pairs of nodes in the KG are not duplicates.



\bibliographystyle{ACM-Reference-Format}
\bibliography{sample-base}

\appendix

\end{document}